%% file: tacl2018v2-template.tex
\documentclass{article} 
\usepackage{times}
\usepackage[acceptedWithA]{tacl2018v2} 

\input{math_commands.tex}

\usepackage{hyperref}
\usepackage{url}
\usepackage{booktabs}
\usepackage{multirow}
\usepackage{subcaption}
\usepackage{graphicx}
\usepackage{multicol}
\usepackage{xcolor}
\usepackage{tablefootnote}
\usepackage{tabularx}

\title{Augmenting Transformers with KNN-Based \\ Composite Memory for Dialog
}
\taclpubformattrue

\author{Angela Fan \\ Facebook AI Research \\ Université de Lorraine\\LORIA  \\ angelafan@fb.com \\  
        \And Claire Gardent \\ CNRS/LORIA  \\ claire.gardent@loria.fr
        \And Chloé Braud \\ CNRS/IRIT \\ chloe.braud@irit.fr
        \And Antoine Bordes \\ Facebook AI Research \\ abordes@fb.com \\}

\newcommand{\human}[1]{\fontsize{8}{8}\textcolor{humancol}{\textnormal{#1}}}
\newcommand{\robot}[1]{\fontsize{8}{8}\textcolor{robotcol}{\textnormal{#1}}}

\newcommand{\fetched}[1]{\fontsize{8}{8}{\textnormal{#1}}}

\definecolor{knolcol}{rgb}{0.0,0.2,0.4}
\definecolor{humancol}{rgb}{0.0,0.2,0.4}
\definecolor{robotcol}{rgb}{0.0,0.0,0.0}
\definecolor{topiccol}{rgb}{0.0,0.0,0.0}

\newcommand{\humanname}[0]{\human{\textbf{Human:}}}
\newcommand{\robotname}[0]{\robot{\textbf{Model:}}}

\begin{document}

\maketitle

\begin{abstract}
Various machine learning tasks can benefit from access to external information of different modalities, such as text and images. Recent work has focused on learning architectures with large memories capable of storing this knowledge. We propose augmenting  generative  Transformer neural networks with KNN-based Information Fetching (KIF) modules. Each KIF module learns a read operation to access fixed external knowledge. We apply these modules to generative dialog modeling, a challenging task where information must be flexibly retrieved and incorporated to maintain the topic and flow of conversation. We demonstrate the effectiveness of our approach by identifying relevant knowledge required for knowledgeable but engaging dialog from Wikipedia, images, and human-written dialog utterances, and show that leveraging this retrieved information improves model performance, measured by automatic and human evaluation.
\end{abstract}

\section{Introduction}

Machine learning approaches to various tasks, such as game-playing or dialog, are often dependent on external information. This information can take multi-modal forms, including structured knowledge bases, free text, and images, and also comes in overwhelmingly large quantities. A pressing challenge is to create models that can identify which specific elements of multiple information sources are relevant in a particular context, and incorporate them into standard architectures on each task. In this work, we focus on human-machine dialog and how to efficiently retrieve external knowledge that is relevant to the dialog. We consider two scenarios and for each scenario, retrieve two types of knowledge: (i) knowledge about similar dialog contexts and (ii) external knowledge used to ground the conversation into real world information. 

Knowledge about similar dialog contexts allows for a hybrid retrieval/generative approach to dialog where the system response is generated based not only on a representation of the current dialog context and of the relevant world knowledge, but also based on a response retrieved from a similar dialog context. The retrieved knowledge can be viewed as providing information about structure and dialog sentences, or utterances: which response is likely given a similar context?

External knowledge is also retrieved to improve the semantic content of the dialog model. In one scenario, Wizard of Wikipedia~\cite{dinan2018wizard}, general topics are provided to crowdworkers, who are asked to have in-depth and specific conversations about these topics by referencing specific Wikipedia sentences as knowledge. In this scenario, external knowledge is retrieved from a pre-selected set of Wikipedia sentences associated with the current dialog topic. Retrieval aims to select the sentence that is most relevant at each step of the dialog and thereby to ground system responses in relevant world knowledge (e.g.\ by referring to Star Wars when talking about science fiction). 

In the other scenario, Engaging ImageChat~\cite{shuster2018engaging}, crowdworkers are provided with images and asked to have a conversation inspired by or about the image. In this case, the retrieved external knowledge is images and their associated dialogs. By retrieving images that are similar to the image being talked about, we aim to enrich system responses with knowledge about what is typically mentioned when describing similar images (e.g.\ when talking about an image with dogs, mentioning their breed).  

Our work on incorporating different types and modalities of knowledge is related to methods that strive to add external memory, such as knowledge bases, to neural networks. Previous work has explored incorporating large external memories into neural network layers \citep{weston2014memory,sukhbaatar2015end,sukhbaatar2019augmenting,lample2019large}. Many existing approaches focus on using attention over the memory slots, which is computationally intensive and becomes less effective as the the size of the memory grows. In this work, we propose representing multiple sources of external information as fixed encodings and using K Nearest Neighbors search to fetch relevant information. KNN search is computationally efficient and scalable, and libraries like \texttt{faiss} \citep{johnson2019billion} allow KNN to be easily used on GPUs and integrated into neural networks. Further, the external memories are pre-encoded, so the information encoding is only computed once. As the external memories are kept fixed, they do not require any training to learn the memories along with the model. We can thus scale easily to larger memories by learning only the KNN-based read operation to identify relevant information from the memory.

Our core contribution proposes an efficient, KNN-based Information Fetching (\textit{KIF}) module that can access relevant external knowledge, combine knowledge from different sources, and integrate this information into standard sequence to sequence architectures. We apply these flexible modules to two dialog datasets that challenge generative models to leverage external information to write coherent, on-topic responses. Both of our chosen tasks require models to leverage external information, such as information from Wikipedia or images, to engage in the conversation. We show that relevant information can be identified from hundreds of thousands of candidates in a multi-modal, multi-knowledge-source setting to improve the performance of generative dialog models. Further, the output of the KIF modules is interpretable as specific human-readable knowledge elements are selected, allowing users to better understand the information the generative model conditions upon when writing the subsequent utterance.
On both datasets, we achieve state-of-the-art results compared to generative models and find there is no statistically significant difference in the interestingness or human preference of our model output compared to state-of-the-art retrieval models.

\section{Related Work}

We discuss related work on learning to incorporate external knowledge into neural networks and efficiently access relevant information. We then describe work in generative dialog that incorporates knowledge.

\begin{figure*}[t]
    \centering
    \includegraphics[width=0.9\linewidth]{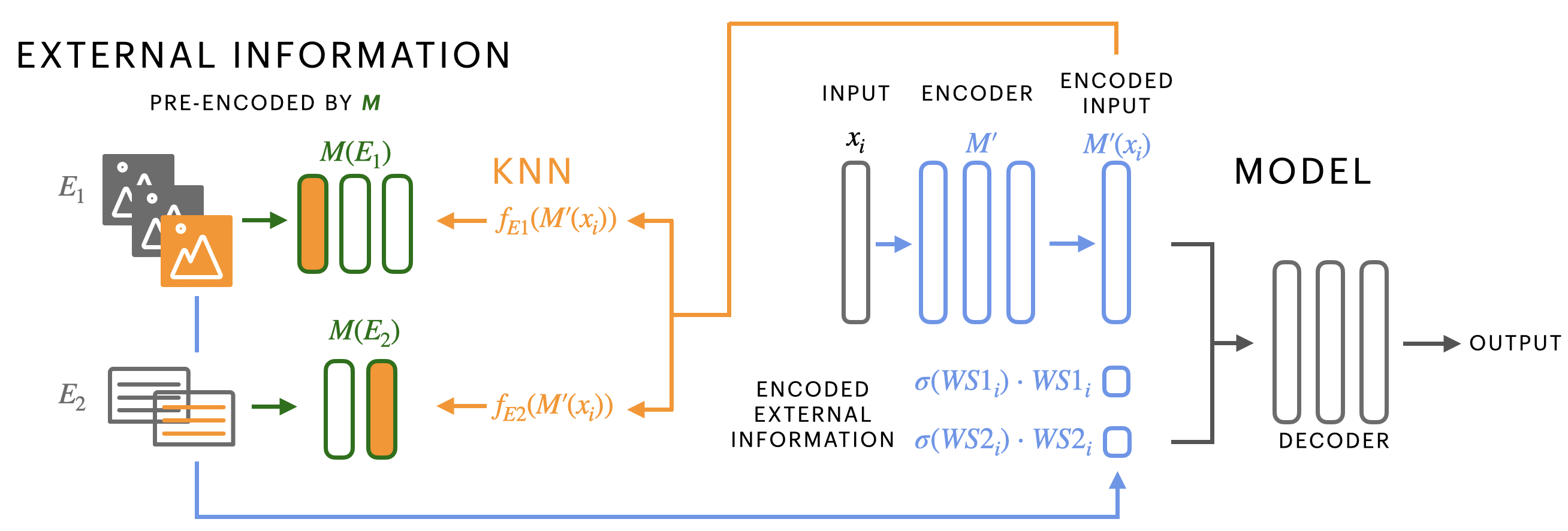}
    \caption{\textbf{KIF} modules fetch relevant information from multi-modal external knowledge. External knowledge sources $E_1$ and $E_2$ are pre-encoded by encoder $M$ (green). In the model, input $x_i$ is encoded by encoder $M'$ (blue) to produce $M'(x_i)$. KIF modules (orange) operate on $M'(x_i)$ and identify the nearest neighbors encoded in $M(E_1)$ and $M(E_2)$ using KNN. Identified relevant elements from $E_1$ and $E_2$ are re-encoded by $M'$ in a gating mechanism with a weighted sum (represented by $\sigmoid(\texttt{WS1}_i) \cdot \texttt{WS1}_i$, where $\texttt{WS}$ stands for weighted sum), then concatenated to $M'(x_i)$. Full description with notation can be found in Section 3.}
    \label{fig:model_general}
\end{figure*}

\subsection{Incorporating External Knowledge}

Augmenting neural networks with memory, or longer term components that can be accessed with read and write operations, has been explored in various proposed architectures. For example, Memory Networks \citep{weston2014memory,sukhbaatar2015end,sukhbaatar2019augmenting} introduce attention mechanisms over large external memories. Neural cache models \citep{grave2016improving} simplify these to access previous memories with a dot product. 
Previous work has also studied how to read and write into these memory architectures  \citep{rae2016scaling,graves2014neural,joulin2015inferring}. In contrast, we focus on how to read large memories.

Another line of research has focused on computational scalability for larger external memories to allow efficient access of information. For example, \citet{chandar2016hierarchical} propose a hierarchical memory network rather than a flat one and \citet{rae2016scaling} learn sparse operations to read and write. \citet{lample2019large} focus on learning memories of up to one million slots and how to efficiently access the slots using product keys. \citet{khandelwal2019generalization} use nearest neighbor operations to augment language models by performing retrieval at the token level --- in contrast, we focus on multi-modal retrieval of multiple pieces of knowledge based on an entire dialog context. Beyond explicit memory representations, it may be possible to store information implicitly during training time by memorizing common patterns present in text \citep{petroni2019language}. We focus on learning to fetch relevant information from multiple explicit external multi-modal knowledge sources and integrate them into one network. Further, our work allows the retrieved information to be interpreted as each memory slot is an explicit fact that can be read as text, rather than a learned vector such as in \citet{lample2019large}.

Work has also focused on computationally efficient softmax operations \citep{mnih2009scalable,grave2017efficient,chen2016strategies}. Many approximate softmax techniques use KNN-like operations to form clusters, and the overall softmax operation is constrained by the slow calculation of the exponential. Our usage of KNN benefits from efficient and scalable libraries such as \texttt{faiss} and \texttt{nmslib}. 

\subsection{Generative Dialog} 

We develop a general architecture for incorporating external information and apply it to the case of generative dialog models. Previous work in dialog has leveraged knowledge as necessary information to accomplish the task. For example, airline and restaurant booking tasks often use API calls to access information about reservation times and availability \citep{bordes2016learning}. In contrast, our work focuses on how to incorporate unstructured knowledge, such as free text found on the web. Previous work has employed architectures that attend over the available knowledge and identify relevant pieces of information, which scales poorly with large quantities of information \citep{dinan2018wizard,qin2019conversing,lian2019learning}. We replace the use of attention over external information with the output of a KNN module. Other work has investigated incorporating information retrieval in language modeling and question answering \cite{chen2017reading,fan2019using,seo2019real,guu2020realm}, while we focus on dialog applications and flexibly incorporating knowledge from multiple, multi-modal sources.

On the modeling side, work has explored both generative \citep{serban2016building,serban2016generative} and retrieval based models \citep{zhang2018personalizing}, which identify the best utterance from the training set to return as the dialog response. This often leverages self-attention or cross-attention mechanisms \citep{humeau2019real}. Further work has explored hybrid models, for example using the output of a retrieval model as input for a generative model \citep{dinan2018wizard,weston2018retrieve,cai2019retrieval,zhu2020reboost}. Some of this work has specialized to use both types of models to generate conversations in an ensemble~\cite{song2016two} or to specifically improve consistency~\cite{song2020generate}. 
We extend these approaches by augmenting generative models with retrieval-like operations based on KNN search, allowing dialog models to flexibly incorporate various sources of external knowledge at the same time and scale to large quantities of retrieval candidates.

\section{KNN-based Information Fetching Modules}

Broadly, the KNN-based Information Fetching (KIF) module assumes an encoder model $M$ can access inputs $\bm{X} = \{ x_1, x_2, \dots, x_n \}$. For example, $\bm X$ can be a collection of sentences, and $x_i$ represents an individual sentence. In a setting without additional supporting information, the encoder will process an input $x_i$ and produce the encoder output $M(x_i)$. 
If $x_i$ is a sequence such as a sentence, then $M(x_i)$ is a representation of the variable size of the sequence length by the fixed size encoder $M$'s hidden size. 
However, in many tasks, additional information is present, represented as $\bm{E} = \{ e_1, e_2, \dots, e_m \}$. We encode each element of $\bm{X}$ and $\bm{E}$ into a vector representation using the encoder. To identify the closest information in $\bm E$ that is relevant to $x_i$, our general approach will be to use K Nearest Neighbors by comparing the representation of $x_i$ with the representation of each element in the set $\bm E$. K Nearest Neighbors is a fully differentiable operation~\citep{plotz2018neural}, so can be incorporated in a straightforward way into neural models. The most relevant information in $\bm E$ will then be available in the model.
We display a KIF-Augmented model in Figure~\ref{fig:model_general} and describe how the KIF module operates.

One challenge to overcome is that the representation of all elements of the knowledge source $\bm E$ are pre-computed and kept fixed, creating $M(\bm E)$ --- we do not backpropagate to affect the embeddings of the pre-encoded knowledge. In the early stages of training, the model receives large amounts of loss, which would affect the quality of the pre-encoded embeddings if we backpropagated to them. Further, encoding the fixed external knowledge once and re-using it allows for greater scalability.
However, this lack of backpropagation can introduce a mismatch between the encoding of $\bm E$ and the encodings produced by a model that is training, as the training model has constantly changing representations because the weights are being learned. We use $M$ to represent the original encoder model used to encode $\bm E$ and $M'$ to represent the constantly training model that is encoding $\bm X$.
The model must learn a function to align $M'(x_i)$ to the pre-encoded elements of the external memory $M(\bm E)$.

To circumvent this misalignment, we learn a mapping operator $f_E(M'(x_i))$ that trains to map elements of the model's representation of $\bm X$, or $M'(\bm X)$, into the additional information representation space $M(\bm E)$.
Concretely, $f_E(M'(x_i))$ is a multi-layer perceptron with ReLU nonlinearities.  From the input  elements of $\bm X$, $f_E(M'(x_i))$ learns representations of an  output close to the corresponding projection of $\bm X$ into $\bm E$. 
This can be interpreted as learning a read operation on a fixed external memory.  If there was no change to the encoding of the model compared to the pre-computed knowledge, then the ideal mapping operator would be the identity function (as $M'$ would equal $M$). However, as the model changes significantly during the training process, the nonlinear mapping capability of $f_E(M'(x_i))$ is essential to be able to identify the correct knowledge $\bm E$ from the input $\bm X$. 

Thus, a model augmented with KIF will incorporate external knowledge in the following manner.
First, we find the $k$ nearest elements to $f_E(M'(x_i))$ in $M(\bm {E})$, based on KNN search with inner product. Then, the relevant elements identified by KNN are re-encoded by $M'$. For example, if element $e_j$ is retrieved by KIF, it would produce $M'(e_j)$.
We use the optimized \texttt{faiss} library for KNN search, which can conduct billion-scale KNN efficiently on GPUs. 

The KNN output for an element $x_i$ is produced by using \texttt{faiss} to search for the $k$ nearest representations to  $f_E(M'(x_i))$ in $M(\bm{E})$. Note that as the encoders $M$ and $M'$ produce output representations of variable length (for example, in the case where $x_i$ is a variable length sequence, such as a sentence), we average across the length dimension to produce a fixed-size representations $r$ to conduct the KNN search.
\begin{align}
    r_{x_i} &= \texttt{Avg}\big(f_E(M'(x_i))\big) \\
    \bm{R}_{E} &= \big\{ \texttt{Avg}(M(e)) \mid e \in \bm{E} \big\} \\
    \texttt{KNN}_{x_i} &= \texttt{KNearest}\big(k, r_{x_i}, \bm{R}_{E} \big) 
\end{align}
Then, the KIF module output for an element $x_i$ is the set of all re-encoded representations of the KNN-retrieved knowledge:
\begin{align}
    \texttt{KIF}_{x_i} &= \big\{M'(e) \mid e \in \texttt{KNN}_i \big\} \label{eqn:kif}
\end{align}
These elements are weighted by their normalized nearest neighbor scores and then summed. 
This is subsequently concatenated to $M'(x_i)$ to form the final encoder output:
\begin{gather}
    [M'(x_i), \texttt{WeightedSum}(\texttt{KIF}_i)]
\end{gather}

This can be easily extended to using multiple modules simultaneously. For instance, two sources of external information, $\bm{E}_1$ and $\bm{E}_2$, can be combined by identifying the top candidates of each information source.
The weighted sum of the KIF output on each information source is concatenated with the encoded input $M'(x_i)$. The KIF output dimensionality is the same size as the hidden size of $M'(x_i)$, so they can be directly concatenated.  

Finally, different sources of information may not be required for every prediction and some information sources can be more important than others. To allow the model to make more fine-grained decisions about what information to use from what source, and how much of it, we add a gating mechanism using a sigmoid function around each weighted sum of KNN representations.
$\texttt{KIF1}_i$ and $\texttt{KIF2}_i$ denote the KIF module from Equation~\ref{eqn:kif} applied to $\bm{E}_1$ and $\bm{E}_2$ respectively.
\begin{gather}
\texttt{WS1}_i = \texttt{WeightedSum}(\texttt{KIF1}_{i}) \\
\texttt{WS2}_i = \texttt{WeightedSum}(\texttt{KIF2}_{i}) 
\end{gather}
which produces the final encoder output, a concatenation of $M'(x_i)$ with the output of multiple KIF modules:
\begin{gather}
\big[M'(x_i),~~\sigmoid(\texttt{WS1}_i) \cdot \texttt{WS1}_i,~~\sigmoid(\texttt{WS2}_i) \cdot \texttt{WS2}_i \big] \label{eqn:gating}
\end{gather}

This concatenation represents the output of the encoder $M'$ and can be used for various purposes, such as providing the encoder output to a decoder in a sequence to sequence model. 

\section{Applying KIF to Dialog Tasks}

We describe how to apply KIF to the task of generative dialog, a  setting where models must generate engaging and on-topic responses. We investigate dialog for two reasons: first, dialog agents must be able to consult relevant information to maintain the topic of the conversation. Second, retrieval-based agents have strong performance compared to generative ones, due to their ability to copy dialog utterances from the training set. Using KIF, we can incorporate the benefits of retrieval architectures into generative, knowledge-based models. 

\subsection{KIF for Generative Dialog} In dialog, $x_i$ represents the text of the conversation $i$. 
A conversation consists of multiple back-and-forth \textit{utterances} (or turns).
For example, a conversation could consist of 4 turns: $x_i = [x_{i,1}, x_{i,2}, x_{i,3}, x_{i,4}]$ where $x_{i,4}$ is the direct utterance the model should respond to, and the earlier utterances are the \textit{conversation context}.

Standard generative dialog models use a Transformer neural network as the encoder $M$ and want to produce an output that is an appropriate response to the conversation.
However, in many cases, the conversation history alone does not include all of the information required to produce an appropriate response.
For example, if a model needs to chat about a specific movie, it can be helpful to provide the model with more information about that movie so a more interesting dialog response could be produced.
To incorporate knowledge, models often concatenate a  knowledge source $\bm E$ such as Wikipedia to $x_i$ and use attention modules to identify the most relevant knowledge. 
However, this approach is computationally intensive when handling large quantities of information.
Further, attention mechanisms have been found to operate poorly over long sequences, as the mechanism becomes blurry due to the softmax and struggles to make fine-grained decisions \citep{fan2018hierarchical}. The same is true for hierarchical approaches, which lack scalability.

We augment Transformer sequence to sequence (seq2seq) networks on the encoder side with KIF to improve generative dialog models.  We experiment on two dialog tasks, Wizard of Wikipedia \citep{dinan2018wizard} and Engaging ImageChat \citep{shuster2018engaging}. In both  datasets, models must leverage information external to the dialog history alone --- in Wizard of Wikipedia, the chat requires access to knowledgeable facts and in Engaging ImageChat, discussion about a specific image. As models must process multiple inputs and ground responses in the knowledgeable facts or images, these tasks challenge existing seq2seq approaches.

\subsection{Wizard of Wikipedia}

The goal of the Wizard of Wikipedia dataset is to train knowledgeable agents that can chat in any domain. The dataset contains 1,365 various topics discussed in 18,430 dialogs in the training set, totalling 166,787 training utterances. Each topic is a general concept, such as \textit{dogs} or \textit{ice cream}, and is included as the first utterance of the conversation. The conversation is meant to be in-depth and detailed, so individual utterances must reference specific knowledge as a basis for the utterance.
The knowledge takes the form of Wikipedia sentences. For example, the chat utterance \textit{I love Toy Story! It was released in 1995} would reference the Wikipedia sentence \textit{Toy Story is a 1995 American computer-animated buddy comedy [...]}. For each utterance, a set of sentences are identified by an information retrieval system, and the crowdworker selected one knowledge sentence as the basis for their utterance. 

\paragraph{Knowledge Sources.} 

Our model for Wizard of Wikipedia has access to two sources of external information, $\bm{E}_1$ and $\bm{E}_2$: 
\begin{itemize}\setlength{\itemsep}{-2pt}
    \item \textit{$\bm{E}_1$ is Wikipedia Knowledge} provided by the dataset as evidence to support knowledgeable chitchat (initially curated by the information retrieval system used in ~\citet{dinan2018wizard}). The scale of this KNN search is to filter through an average of 34 sentences.  The KIF module uses dialog features to fetch relevant knowledge to condition upon to generate the subsequent utterance. 
    \item \textit{$\bm{E}_2$ is Training Utterances}. To incorporate the benefits of retrieval-based dialog models to the generative setting, we use KIF to identify relevant utterances from the training set and take their \textit{responses} as input. If many conversations about dogs have already occurred, models should be able to take advantage of these human-written examples to improve their generations. For example, likely conversation could occur about the breed of the dog, daily routine with a pet, and similar topics. There are around $170$K dialog utterances as inputs to KNN search.  This can be interpreted as incorporating the benefits of retrieval models by identifying an utterance with similar structure as the text the model would like to generate. We do not allow the module to fetch the correct response of the current conversation context.
\end{itemize}

Access to these two sources of knowledge can be seen as learning a template and a topic separately. Sample templates can be identified from the training utterances, and topic-specific information learned by accessing the Wikipedia knowledge. 

\paragraph{Additional KNN Features.} 
To better identify relevant training utterances from the large quantity available, we break down $x_i$ into conversation sub-features for a more fine-grained match in the KNN search step. By conducting KNN on more features, we can achieve higher quality retrieval. We leverage the nature of dialog to decide these features.

We concatenate the encoding of the most recent dialog utterance (e.g.\ $x_{i, \mathrm{last}}$) with the encoding of the dialog context from the current conversation and the turn number $t$, such that
$M'(x_{i, \mathrm{last}}), M'(x_{i, \mathrm{-last}}), t$
is the representation used for KNN search. Concretely, if the model is trying to produce the 5th turn of the conversation, then $x_{i, \mathrm{last}}$ is the most recent utterance from the dialog partner, $x_{i, \mathrm{-last}}$ would be the last 3 turns of exchange, and $t$ would be 4.
Note that the turn number is represented as a standalone number.
These are known to be salient conversation features. The most recent dialog utterance is the direct turn the model is responding to, and the dialog context may provide additional clues. The turn number is important, as earlier turns are often  generic (e.g.\ \textit{how are you doing today}) and later turns are more specific.

\subsection{Engaging ImageChat}

The goal of Engaging ImageChat is to create agents capable of chitchatting about images selected from the YFFC100M dataset \citep{thomee2016yfcc100m}. The dataset contains 186,782 dialogs in the training set, each about a unique image, totalling 355,862 utterances. Agents are assigned one of 215 personalities (e.g.\ \textit{sweet, caring, excited}) to increase engagingness. Previous work~\cite{shuster2018engaging,shuster2019engaging} identified that both crowdworkers and models, when provided with personalities, produced more diverse, interesting responses, as evaluated by humans.

We use a Multi-Modal neural network designed to handle both image input and text input. Following \citet{shuster2018engaging},  the images are encoded using a pre-trained ResNeXt network \citep{xie2017aggregated}. To extract the final image representation, we project the $2048$-dimensional output of the image encoder to $512$-dimensions using a deep multi-layer perceptron with ReLU activation units. The conversation history, which includes the one-word personality, is encoded with a Transformer encoder network. The image and conversation are integrated using the Multimodal-Sum-Combiner module proposed in \citet{shuster2018engaging}.

\begin{table*}
\centering
\begin{tabular}[t]{lrr}
\toprule
Model & Test F1 & Test F1 \\
 & (Seen) & (Unseen) \\
\midrule
\bf Retrieval Baselines & & \\ 
Retrieval Transformer MemNet \cite{dinan2018wizard} & 15.4 & 12.4 \\ 
\midrule
\bf Generative Baselines & & \\ 
2-Stage Generative MemNet \cite{dinan2018wizard} & 18.9 & 17.4 \\ 
Generative Transformer MemNet \cite{dinan2018wizard} & 16.9 & 14.4 \\ 
\hspace{1cm} + Reddit Pre-Training & 17.6 & 16.3 \\ 
Retrieve and Refine \cite{weston2018retrieve}& 18.2 & 17.9 \\ 
Response Generation with MR \cite{qin2019conversing} & 17.5 & 16.8 \\ 
\midrule 
KIF-Augmented Transformer & \bf 25.9 & \bf 22.3 \\ 
\bottomrule
\end{tabular}
\caption{Results on the \textbf{Wizard of Wikipedia} dataset. We implement the Retrieve and Refine and Response Generation with MR approaches, all with Reddit Pre-Training, and evaluate them on Wizard of Wikipedia. The \textit{Seen} test set consists of conversations on topics seen at training time, and the \textit{Unseen} test set consists of conversations about new topics that were not in the training set.}
\label{tab:wizard_results}
\end{table*}

\paragraph{Knowledge Sources.} 

Our model for Engaging ImageChat has access to two sources of external information, $\bm{E}_1$ and $\bm{E}_2$:
\begin{itemize}\setlength{\itemsep}{-2pt}
    \item \textit{$\bm{E}_1$ is Chat on Similar Images}. While there are over $180$K different images in this dataset, many of the images are similar. For example, conversations associated with two pictures of dogs could be relevant to each other. The model is able to use KIF directly on the current image features to fetch from around $180$K different images and return 6 turns of related chat for each fetched image. Fetching from $\bm{E}_1$ consists of identifying related image chats, or conversations on related topics. 
    \item \textit{$\bm{E}_2$ is Training Utterances}. Similar to the motivation for the previous dataset, we allow the model to identify training utterances that could be useful for responding in the current conversation. The scale of this fetching task is large: $350$K dialog utterances. This could be interpreted as identifying utterances with similar structure to what the model would like to generate, and is complementary to the topic-based related image chats. 
\end{itemize} 

\paragraph{Additional KNN Features.} 

To identify relevant information from training utterances, we use the same dialog features as Wizard of Wikipedia in the KNN search step, with one modification: we add the \textit{personality} provided by the dataset. We represent the personality feature as the personality word, such as \textit{caring}, and embed it with the encoder $M'$. As utterances from speakers with the same personality are more likely to be similar, this feature improves the quality of the fetched information. For example, conversations with the \textit{sweet} personality often include similar text such as \textit{aww, that's wonderful}. We use two additional features for the KNN search: $t$, the turn number, and $p$, the personality. This feature is explicitly used in ~\citet{shuster2018engaging} to improve the engagingness and flow of the conversation. Similar to Wizard of Wikipedia, we represent the conversation turn $t$ as a number. The Transformer model is used to encode text $x_i$ and produce a representation of the text, then the turn number $t$ and personality $p$ are represented separately. As the personality is a word, we use the same Transformer to encode it. The concatenation of features used for KNN search is: $M'(x_{i, \mathrm{last}}), M'(x_{i, \mathrm{-last}}), t, p$.

\begin{table*}
\centering
\begin{tabular}[t]{lr}
\toprule
Model & Test F1 \\
\midrule
\bf Retrieval Baselines & \\ 
Retrieval Transformer \cite{shuster2018engaging} & 9.8\tablefootnote{In \citet{shuster2018engaging}, retrieval Transformer models report Hits@N using a fixed candidate set of 99 distractor candidates and 1 true candidate. We compute F1 using their open-sourced model by scoring the entire training set of over $350$K utterances with the model and taking the top scoring candidate as the response.} \\  
\midrule
\bf Generative Baselines & \\ 
Generative Transformer MemNet \cite{dinan2018wizard} & 7.1 \\ 
\hspace{1cm} + Reddit Pre-Training & 12.8 \\ 
Retrieve and Refine\cite{weston2018retrieve}  & 13.6 \\ 
Response Generation with MR \cite{qin2019conversing} & 13.2 \\ 
\midrule 
KIF-Augmented Transformer & \bf 14.4 \\ 
\bottomrule
\end{tabular}
\caption{Results on the \textbf{Engaging ImageChat} dataset. We implement the Generative Transformer Memory Network, Retrieve and Refine, and Response Generation with MR approaches, all with Reddit Pre-Training, and evaluate them on Engaging ImageChat.}
\label{tab:imagechat_results}
\end{table*}

\section{Experimental Setup}

\subsection{Implementation Details} 

\paragraph{Parameter Settings.} 
We use \texttt{parl.ai} \citep{miller2017parlai} to implement our models. The data for both datasets used is available for download from \texttt{parl.ai} as well. We use byte-pair encoding \citep{sennrich2016neural} to represent the text to better handle the rare word problem  \citep{dinan2018wizard,fan2018controllable}.  Our generative Transformer models have 8 encoder layers and 8 decoder layers, with FFN size 2048, embedding dimension 512, and 4 attention heads. We optimize using Adam \citep{kingma2014adam} and the inverse square root learning schedule \citep{vaswani2017attention} with 10k warmup updates. The initial learning rate is 0.0001 and we optimize for model perplexity.  We use a dropout of 0.5 and set gradient clipping to 0.1.  We set k = $5$ for all cases. For both datasets, we model a vocabulary size of 54944 based on the BPE-based vocabulary from the Reddit pre-training. We tuned the learning rate and batchsize hyperparameters together. 

\paragraph{Pre-training.} We pre-train the Transformer seq2seq model used for both datasets on 250M comments from Reddit. The Reddit dataset was made available by \texttt{pushshift.io}. The comments are parsed to maintain conversational threads of users responding to each other, so the encoder network has been exposed to conversational context at training time. Note that the Reddit dataset does not include aspects such as personality, as those are unique to specific datasets such as Engaging ImageChat. The context size in pre-training is set to 512 tokens.
The ResNeXt encoder used to model images for the Engaging ImageChat dataset was pre-trained on 3.5 billion images \citep{mahajan2018exploring}.

\subsection{Evaluation}

\paragraph{Generation.} We generate with beam search, setting the beam size to $4$. We use 3-gram blocking. This technique disallows repeated n-grams from being generated multiple times and reduces repetition.

\begin{figure*}[t!]
    \centering
    \includegraphics[width=.85\linewidth]{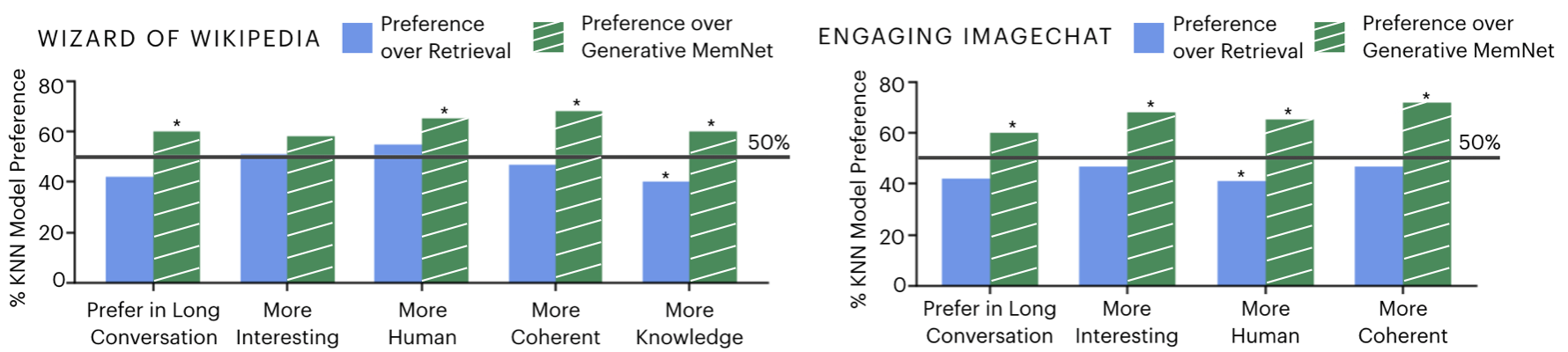}
    \caption{\textbf{Human Evaluation Results on both Datasets}. More than 50\% indicates the KNN Model is preferred. Stars indicate statistical significance at $p < 0.05$.}
    \label{fig:human_wizard}
\end{figure*}

\begin{figure}[t!]
    \centering
    \includegraphics[width=0.85\linewidth]{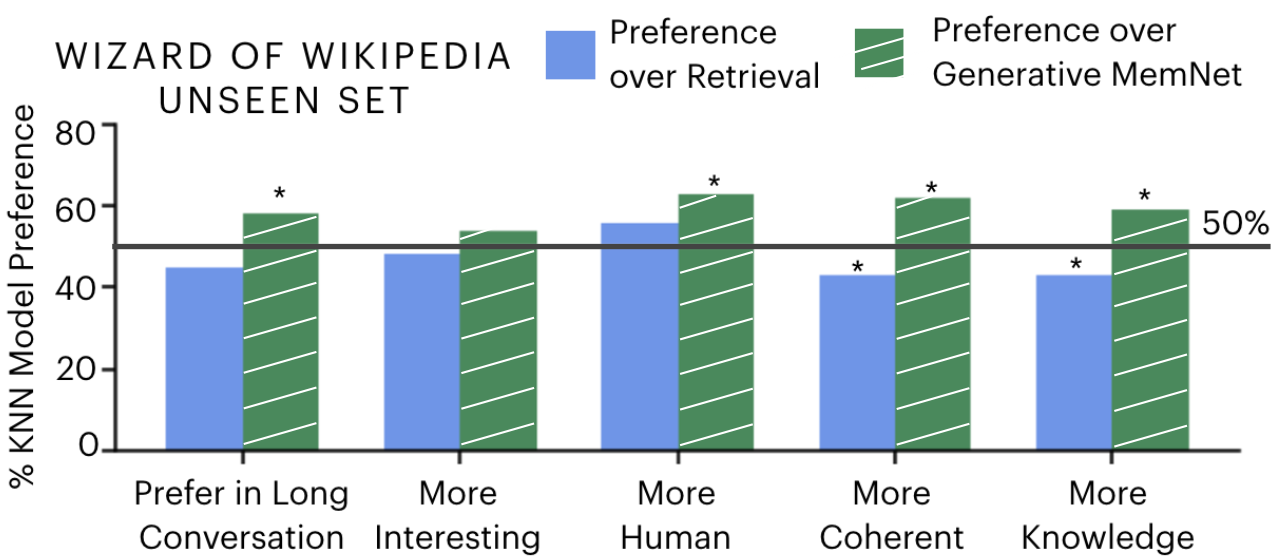}
    \caption{\textbf{Human Evaluation on the Unseen Test set} of Wizard of Wikipedia. More than 50\% indicates the KNN Model is preferred. Stars indicate statistical significance at $p < 0.05$.}
    \label{fig:human_wizard_unseen}
\end{figure}

\begin{figure*}[ht]
\small
\centering
\def\arraystretch{0.92}
\setlength{\tabcolsep}{2pt}
\begin{tabular}{r p{12cm}}
\toprule
\humanname & \human{I love watching disney movies! Do you like Disney?} \\
\robotname & \robot{I love disney, I love watching disney movies and different animations and characters that make them unique} \\
    & \fetched{\textbf{Fetched Knowledge:} \textit{disney consumer products and interactive media - the company  specializes in producing animated films}} \\ 
    & \fetched{\textbf{Fetched Training Utterance:} \textit{i love music, I love going to music shows and dancing and having fun}} \\ 
\humanname & \human{What's your favorite disney movie? There are so many good ones.} \\
\robotname & \robot{I love the incredibles, they are my favorite disney movie.}\\
    & \fetched{\textbf{Fetched Knowledge:} \textit{disney announced intentions to develop additional superhero films after the success of the incredibles}} \\ 
    & \fetched{\textbf{Fetched Training Utterance:} \textit{i love kiteboarding, it is one of my favorite activities on the water.}} \\ 
\humanname & \human{omg, me too! the second one just came out i think. I don't watch a lot of sequels though, I think the last one I saw was toy story.} \\
\robotname & \robot{That's cool! I remember when toy story was first released in 1995.} \\
    & \fetched{\textbf{Fetched Knowledge:} \textit{talk of a second film following toy story began around a month after the film's opening, in december 1995}} \\ 
    & \fetched{\textbf{Fetched Training Utterance:} \textit{that's neat! i like so many of her songs! i remember when her first album was released in 2008.}} \\ 
\humanname & \human{I've actually never seen toy story before. Kind of embarrassing...} \\
\robotname & \robot{Ohhh ... I think toy story is a classic? It was first released in 1999.} \\
    & \fetched{\textbf{Fetched Knowledge:} \textit{toy story 2 is a 1999 american computer-animated comedy film produced by pixar animation studios for walt disney pictures}} \\ 
    & \fetched{\textbf{Fetched Training Utterance:} \textit{Ohhh ... What do people with color blindness do to cope with the effects?}} \\ 
\bottomrule
\end{tabular}
\caption{\textbf{Conversation between Human and KIF-Augmented Transformer on Wizard of Wikipedia}. The top-scoring Wikipedia knowledge and training utterances fetched by KIF are displayed with model output.}
\label{fig:examples}
\end{figure*}

\paragraph{Automatic Metrics.}

Following \citet{dinan2018wizard}, we compute \textit{F1}, a metric of unigram overlap, between the generated utterance and the human-written reference utterance from the dataset. For generative models, utterances are generated using beam search. For retrieval models, the next utterance is predicted by ranking the entire set of training utterances, and the highest scoring utterance is chosen. 

In Wizard of Wikipedia, there are two test sets: a set of \textit{seen} topics, or topics that have been seen at training time with new test-time dialogs. The second set is \textit{unseen}, or topics that have not been encountered at all during training time. We evaluate on both subsets.

\paragraph{Human Evaluation.} 

We follow the setup and use the analysis questions proposed in the \texttt{Acute-Eval} dialog evaluation system \citep{li2019acute}.
For reproducibility, we adopt this existing evaluation setting that has been applied to several dialog datasets. We use the question wording suggested by \texttt{Acute-Eval} and follow their self-chat procedure and interface. As one of the original datasets assessed in this system was Wizard of Wikipedia, their evaluation setting extends naturally to ours. 
We collect 100 human-bot conversational dialogs on a crowdsourcing platform for both datasets. 
The dialogs are eight turns long.
Then, we show pairs of the collected conversations side by side, one conversation with a human and model A and the other conversation with a human and model B. We ask annotators the following questions:
\begin{itemize}\setlength{\itemsep}{-2pt}
    \item Who would you prefer to talk to for a long conversation? 
    \item If you had to say one of the speakers is interesting and one is boring, who would you say is more interesting?
    \item Which speaker sounds more human?
    \item Which speaker has more coherent responses in the conversation? 
    \item If you had to say that one speaker is more knowledgeable and one is more ignorant, who is more knowledgeable? (Wizard of Wikipedia only)
\end{itemize}

We measure the percentage of time one model was chosen over the other, taking the majority agreement between three evaluators. To reduce variance, dialogs paired in the evaluation were collected on the same topic for Wizard of Wikipedia and collected on the same image and personalities for Engaging ImageChat. Topic and images selected for evaluation are unique and taken randomly from the test set.

\begin{figure*}[!htbp]
\small
\centering
\def\arraystretch{0.92}
\setlength{\tabcolsep}{2pt}
\begin{tabular}{r p{10cm}}
\toprule
\humanname & \human{Hey, how are you doing} \\
\fetched{\textbf{Fetched Training Utterances:}}  & \fetched{\textit{I'm great, thanks for asking. Craving some chocolate. Do you like chocolate?}} \\ 
    & \fetched{\textit{Hello, how is it going? I know some trivia about this movie}} \\ 
    & \fetched{\textit{Hello, it’s lunch time here, and I’m in the mood for a great steak}} \\ 
\fetched{\textbf{Model Response:}} & \fetched{Hello, I'm great, thanks for asking}   \\ 
    & \\ 
\humanname & \human{What are your hobbies?} \\
\fetched{\textbf{Fetched Training Utterances:}}      & \fetched{\textit{I work at an elementary school. I hope you find a job you love too [...]}} \\ 
    & \fetched{\textit{I have a hound, we just got her. Although, I grew up with Labrador Retrievers.}} \\ 
    & \fetched{\textit{I just love ice cream. I love the types with fruits and flavours. Do you like ice cream?}} \\
\fetched{\textbf{Model Response:}} & \fetched{I work at a tennis court. I like to play tennis. Do you like your hobby?}   \\ 
\midrule
\humanname & \human{hi buddy, what do you think about cinematography?} \\
\fetched{\bf Gold Chosen Knowledge:} & \fetched{\textit{cinematographers use a lens to focus reflected light from objects into a real image [...]}} \\ 
\fetched{\textbf{Fetched Knowledge:}} & \fetched{\textit{cinematography is the art of motion-picture photography}} \\ 
    & \fetched{\textit{typically, a lens is used to repeatedly focus the light reflected from objects [...] }} \\ 
    & \fetched{\textit{the modern photographic camera evolved from the camera obscura}} \\ 
\fetched{\textbf{Model Response:}} & \fetched{I love cinematography, it is my hobby. It is a kind of motion-picture photography.}   \\ 
    & \\ 
\humanname & \human{Speaking of blue skies, have you seen the 1946 movie staring bing crosby?} \\
\fetched{\bf Gold Chosen Knowledge:} & \fetched{\textit{blue skies is a 1946 american musical comedy film [...] and starring bing crosby [...]}} \\ 
\fetched{\textbf{Fetched Knowledge:}}     & \fetched{\textit{blue skies is a 1946 american musical comedy film [...] and starring bing crosby [...]}} \\ 
    & \fetched{\textit{blue skies the band has since broken up}} \\ 
    & \fetched{\textit{blue skies was was composed in 1926 as a last - minute addition to betsy the musical}} \\ 
\fetched{\textbf{Model Response:}} & \fetched{Oh cool, that's an American comedy film starring Bing Crosby. Do you like watching movies?}   \\ 
\bottomrule
\end{tabular}
\caption{\textbf{Examples of Top-3 Fetched Training Utterances and Fetched Knowledge} when responding to a human chat from the dataset using a trained Wizard of Wikipedia model. Examples are taken from validation.}
\label{fig:example_beginnings}
\end{figure*}

\subsection{Baselines} 

We compare Transformers augmented with KIF to other existing approaches on Wizard of Wikipedia and Engaging ImageChat. The best approaches, judged by human evaluation, are retrieval models, the Retrieval Transformer Memory Network from ~\cite{dinan2018wizard} and the Retrieval Transformer from ~\citet{shuster2018engaging}. These have been shown to be strong baselines compared to other retrieval techniques based on TF-IDF \cite{chen2017reading}. Thus, we report the existing retrieval models for both datasets, but focus on comparing to other generative baselines.

We compare to three additional generative baselines. Note that in Wizard of Wikipedia, the construction of the dataset is that sentences of Wikipedia knowledge are provided with the utterances in a concatenated form. Models must identify the relevant information in this provided knowledge, or can access more Wikipedia knowledge beyond the provided sentences. The following baseline methods always have access to the information provided in the dataset already, but no additional Wikipedia knowledge beyond that.

\begin{itemize}\setlength{\itemsep}{-2pt}
    \item \textit{Transformer Memory Networks}. To contrast the ability of KIF to existing work, we compare our models to published Transformer Memory Networks \citep{dinan2018wizard}. These models encode each piece of external information independently with a Transformer Encoder, and these are stored as memory slots. To access information in the memory slots, a model performs dot-product attention between the memory slots and the dialog context. In \citet{dinan2018wizard}, the knowledge selection from Wikipedia was supervised with either \textit{(a)} a two-stage model where the first model was trained to predict the right knowledge and a second model conditions on the predicted knowledge to generate the next utterance, or \textit{(b)} an end-to-end model with an auxiliary loss for knowledge prediction accuracy. 
    \item \textit{Retrieve and Refine}. We implement a hybrid model \citep{weston2018retrieve} that incorporates top retrieval candidates as additional input to Generative Transformer MemNets. In Retrieve and Refine, a fixed number of candidates are retrieved and concatenated to the conversational history in the encoder, making the input much longer. For both datasets, the Retrieve and Refine mechanism that fetches a fixed number of training utterances is added to the Generative Transformer MemNet with Reddit Pre-Training baseline. 
    
    Unlike the KIF-Augmented Transformer, the retrieval is conducted with a separate model so there is no backpropagation to affect the retrieval. With KIF, models can alter the retrieved candidates by learning the mapping operator. Further, a fixed amount of information is always retrieved, without the capability to easily rescale to focus on specific candidates. KIF modules have weighting mechanisms to focus more on certain information, and the modules are combined with gating so models can learn which knowledge sources are more important and adjust flexibly. Lastly, Retrieve and Refine is only used to retrieve one source of information: training set utterances. 
    
    \item \textit{Response Generation with MR}. We implement the model proposed in \citet{qin2019conversing}, which encodes the conversation history and document contextually with a biLSTM before generating the next dialog utterance. The initial model was applied to a machine reading task where a knowledge document was provided along with the conversation history. 
    For Wizard of Wikipedia, we replace the knowledge document with the Wikipedia sentences provided in the dataset. The model then uses the conversation to identify the most relevant information in the document using a cross-attention mechanism. For the Engaging ImageChat dataset, as there is no document provided with the dataset, we replace the expected document with the conversation history, and use the most recent utterance in the conversation to attend to the conversation history.
     
    We make an additional improvement to this baseline: in \citet{qin2019conversing}, the embeddings used pre-trained CoVE vectors \cite{mccann2017learned}. We found our Reddit pre-trained Transformer embeddings to work more effectively as they are trained for dialog. Thus, we replace CoVE embeddings with domain-specific ones. 
\end{itemize}

All of Transformer generative baselines are initialized with the same pre-training on Reddit that we use for our models for fair comparison on modeling quality.
  
\section{Results}

We describe the results of incorporating KIF modules into Transformer networks. We display an example conversation between a human and our model in Figure~\ref{fig:examples}, and show the top scoring Wikipedia knowledge and Training Utterance fetched by KIF modules. We compare to various baselines using automatic and human evaluation, and discuss our experiments. We present various ablation settings to understand the key features that make our method function.

\begin{figure*}[ht]
    \centering
    \includegraphics[width=0.85\linewidth]{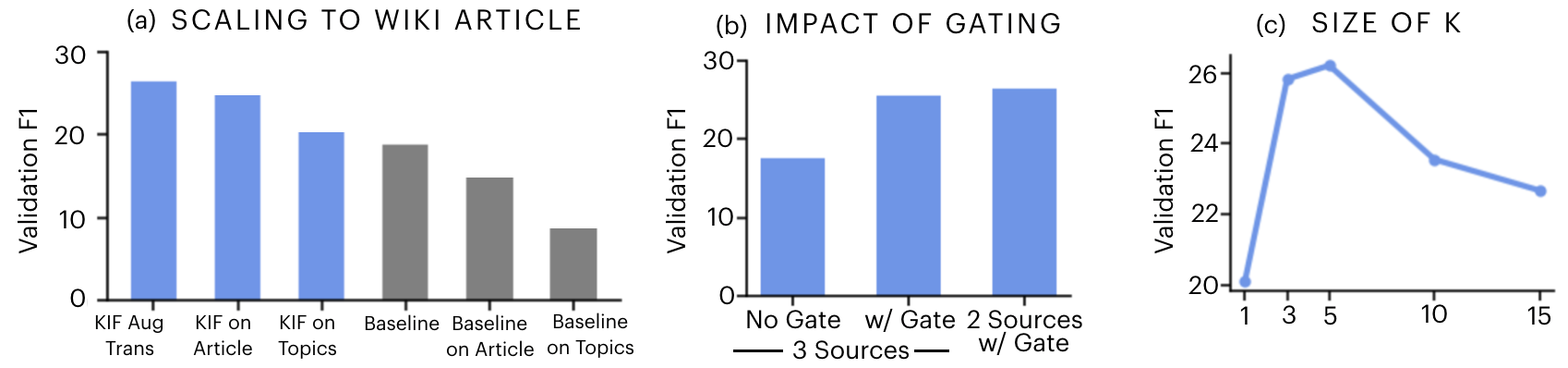}
    \caption{\textbf{Ablations} on Wizard of Wikipedia. (a) KIF can \textbf{scale} to hundreds of relevant sentences  (blue) while the baseline model, the Generative Transformer MemNet (gray), scales poorly (b) \textbf{Gating} can remove irrelevant information. In the 3 Sources case, one source of external information is unrelated. (c) \textbf{Performance as $k$ varies.}}
    \label{fig:ablation_1}
\end{figure*}

\subsection{KIF is Effective for Incorporating Knowledge}

\paragraph{Automatic Evaluation.} Comparing KIF augmented Transformer networks to published baselines and Retrieve and Refine, we find improved results. 

For Wizard of Wikipedia, the improvement in F1 score over the best baseline is around 8 points (see Table~\ref{tab:wizard_results}). A major contributing factor is the construction of the dataset --- as each dialog turn is grounded in a specific knowledge sentence from Wikipedia, improving the ability to identify the relevant fact strongly improves performance. Contrasting the results from the \textit{seen} and \textit{unseen} test sets in Table~\ref{tab:wizard_results}, the improvement on \textit{unseen} is worse --- it is harder to fetch training utterances for unseen topics. 

While Imagechat has no explicit dependency on knowledge, we still see a 2 point improvement compared to the Generative Transformer MemNet (with the additional Reddit pre-training), indicating that KIF can be generally useful (see Table~\ref{tab:imagechat_results}). Compared to an even stronger baseline that we tune in this work, Retrieve and Refine, we see 1 point improvement.

\paragraph{Human Evaluation.} 
Results are shown in Figure~\ref{fig:human_wizard}. On both datasets, we find there is large improvement over existing generative models (green bars) that is statistically significant for some of the evaluation questions. Evaluators agree that KIF-augmented Transformers are generally more coherent and human-sounding compared to the Generative MemNet.

Compared to existing retrieval models (blue) is more nuanced. Along the lines of existing work \citep{zhang2018personalizing,dinan2018wizard}, we find that retrieval-based models score very well in human evaluations that ask how human or interesting a dialog sounds. This is because retrieval models return human-written utterances from the training set and do not suffer from decoding mistakes present in generative models. For example, on Engaging ImageChat, while our model has significantly improved over the generative baseline (see green bars in Figure~\ref{fig:human_wizard}, right), it does not beat retrieval based methods in sounding more human or being more interesting (see blue bars in Figure~\ref{fig:human_wizard}, right). As the Retrieval baseline returns human-written text for other humans to evaluate, we hypothesize that humans score each other's writing quite well. Compared to generative models, which we focus on improving, retrieval models often produce longer text with more interesting, nuanced vocabulary usage, and do not make generation mistakes such as repetition. These factors often lead to the stronger performance of retrieval models. 

A surprising result is that KIF-augmented Transformers are more human sounding than retrieval models on Wizard of Wikipedia. This is because the dataset's utterances are long and factual due to the tendency of crowdworkers to copy Wikipedia. Sometimes humans chatting with the retrieval bot would respond \textit{uh... that's an interesting fact?} Otherwise, our model scores similarly to retrieval models, with most evaluations not having statistically significant difference.

\begin{table*}
    \small
    \centering
    \begin{tabularx}{\textwidth}{XXX}
    \toprule 
         \textbf{Knowledge} & \textbf{Training Utterance} & \textbf{Generation} \\ 
         \midrule
         \textbf{buzz lightyear}'s name is in honor of astronaut edwin `buzz' aldrin & my favorite character in that book \newline series is hermione granger & cool! my favorite character in that movie is \textbf{buzz lightyear} \\ \\ 
         mr potato head is based on the \textbf{real-life mr. potato head} toy & my favorite character in that book \newline series is hermione granger & my favorite character in that movie is \textbf{real-life mr potato head}  \\ \\
         \textbf{slinky dog} is a toy dachschund with a metal slinky for a body & my favorite character in that book \newline series is hermione granger & cool! my favorite character is the \textbf{slinky dog} \\
         \midrule  
         slinky dog is a toy dachschund with a metal slinky for a body & \textbf{i really like} the character hermione granger & cool! \textbf{i really like} slinky dog \\ \\
         slinky dog is a toy dachschund with a metal slinky for a body & \textbf{my favorite character} of all time \textbf{has to be} hermione granger & i love that movie, \textbf{my favorite character has to be} slinky dog the dachshund \\ \\
         slinky dog is a toy dachschund with a metal slinky for a body & i agree with you! that's \textbf{my favorite} character as well & i think so too\textbf{! my favorite} is slinky \\
    \bottomrule 
    \end{tabularx}
    \caption{\textbf{Effect of Fetched Information on Generated Utterances.} The \textbf{top} section provides examples for a fixed training utterance, changing the knowledge --- the generated text maintains the construction of the training utterance but changes the favorite character to match the knowledge. The \textbf{bottom} section provides examples for fixed knowledge but changing the training utterance --- the generated text modifies its form to match the training utterance, but the favorite character information remains consistent.}
    \label{tab:control_examples}
\end{table*}

We conduct a second evaluation on the Unseen Test Set of the Wizard of Wikipedia dataset. Results are shown in Figure~\ref{fig:human_wizard_unseen}. Trends are similar compared to the results on the Seen Test set, though the preference for the KIF-augmented Transformer is greater over the retrieval baseline. We hypothesize that because the Unseen Test Set is on entirely held out topics, the retrieval baseline can struggle to identify relevant utterances. In contrast, the KIF-augmented Transformer, similar to the generative baseline from \citet{dinan2018wizard}, can use the generative capability to produce utterances.

Lastly, we conduct an additional study to examine the variance of the comparative dialog judgements. The evaluation study for Wizard of Wikipedia is repeated three times on different days, and evaluators who have answered on previous days are not allowed to evaluate again in any subsequent experiments. Overall, we find reasonable interannotator agreement rates, around 73\% averaged across all evaluations, which is similar to the agreement rates reported in ~\cite{li2019acute}. We find there is greater variance on questions asking which dialog is \textit{more human} and \textit{more interesting}, most likely as different evaluators can interpret these in different ways. Further, we see that comparison with the Retrieval model has less variance compared to the Generative model, possibly because the Retrieval model's human written text is devoid of mistakes. Overall, we find that the conclusions (and statistical significance) are stable across multiple evaluations.

\subsection{Analysis of Fetched Knowledge}

Example conversations from our KIF-augmented generative model are shown in Figure~\ref{fig:examples} on Wizard of Wikipedia. We find that relevant knowledge is identified that affects the \textit{content} of the generated utterance. For example, the model finds knowledge sentences about Disney movies as the human conversationalist starts the conversation discussing Disney. The model leverages the fetched knowledge to write the content of the generated utterance. In a concrete example, the fetched sentence \textit{disney announced intentions [...] after the success of the incredibles} leads the model to generate the utterance \textit{i love the incredibles, they are my favorite disney movie}. 

In contrast, the model uses the form of the fetched training utterance often as a template for writing a response. For example, the model copies the training utterance \textit{Ohhh ... what do people with color blindness do to cope with the effects?} and starts the model generation with \textit{Ohhh ...} and continues with the question \textit{i think toy story is a classic?} following the form of the selected training utterance.

Figure~\ref{fig:example_beginnings} displays the top-3 fetched training set utterances and knowledge sentences on the Wizard of Wikipedia dataset when responding to a human utterance. KIF modules can identify multiple relevant items. In response to the human question about \textit{blue skies the 1946 movie} the model identifies both the comedy film and the band. 

Finally, the elements retrieved by KIF modules provide a more interpretable understanding of what the model is conditioning upon to generate a dialog response. In Table~\ref{tab:control_examples}, we display for the same dialog history, changing the model's fetched training utterance and knowledge sentence for our own examples. The model heavily incorporates our manual changes of the fetched information into the generated utterance. For example, changing the knowledge directly affects what the model generates as the favorite character --- from \textit{buzz lightyear} to \textit{mr potato head} to \textit{slinky dog} --- while changing the fetched training utterance changes the form of the generated sentence.

\subsection{Scaling KIF to Challenging Retrieval Settings}

KIF modules can be used in more realistic and challenging settings for knowledge retrieval  that test the scalability of the module. 
In Figure~\ref{fig:ablation_1}(a), we compare the Generative Transformer MemNet Baseline with KIF-Augmented Transformers in three settings. The first is the standard Wikipedia sentences provided by the dataset (average 34 sentences). Then, we extend to providing the model with the full Wikipedia article (on average, 57 sentences) and finally to multiple Wikipedia articles (on average, totaling 205 sentences), identified using the conversation's topic.  This increasing size of available knowledge could be realistic for settings where it is unclear what information is most relevant, if filtering steps to preprocess the data remove potentially relevant information, or if information synthesis from multiple knowledge sources is necessary to produce a high quality generation.  As the Wikipedia knowledge becomes more difficult to identify, performance decreases, but still outperforms the baseline that uses the dataset-provided set of 34 sentences. 

Comparing the scaling capability of KIF to the standard Generative Transformer MemNet Baseline highlights the advantage of using KNN. The attention-based mechanism used in \citet{dinan2018wizard} struggles to identify salient information when given increasingly larger quantities of knowledge, unlike the KNN information fetch. We hypothesize the attention mechanism is challenged by softmax-ing over a larger quantity of inputs, as it can be difficult to make sharp distinctions.

\subsection{Ablations}

\paragraph{Importance of Multiple Knowledge Sources.}
\label{sect:multiple_knowledge}

One benefit of the KIF module approach is that several modules can be combined, each capturing information from a different source. In
both settings, Wizard of Wikipedia and Engaging ImageChat, two modules were
used to incorporate multiple forms of knowledge --- training utterances to capture the capability of a retrieval-based model and knowledge from Wikipedia or related chats based on image features.  We perform here an ablation study to evaluate the impact of using only one source of information. As can be seen in Table 4, performance decreases when only one source of information is used (see Table~\ref{tab:no_know_baselines}).

For Engaging ImageChat, this study also underlines the importance of being able to fetch in a multi-modal fashion. The general form of the KIF module --- requiring only a feature vector to find nearest neighbors from --- allows fetching on multiple modalities such as text and images. In Table~\ref{tab:no_know_baselines}, using the Image-based KIF to fetch text from Related Images is important to reach the strongest performance (compare Training Utterances Only that uses text-based KIF and using both Training Utterances and Related Images). 

\begin{table}[t]
\centering
\begin{tabular}[t]{lr}
\toprule
Model & Test F1 \\
\midrule
\textit{Wizard of Wikipedia} & \\ 
Training Utterances Only & 18.1 \\ 
Wiki Knowledge Only & 23.9 \\
Training Utterances and Wiki Knowledge & 25.9 \\ 
\midrule 
\textit{Engaging ImageChat}  & \\ 
Training Utterances Only & 13.9 \\ 
Related Images Only & 13.8 \\ 
Training Utterances and Related Images & 14.4 \\ 
\bottomrule
\end{tabular}
\caption{\textbf{Using Multiple KIF Modules on Multiple Sources} is important for improved performance. 
}
\label{tab:no_know_baselines}
\end{table}

\begin{table}[t]
\centering
\begin{tabular}[t]{lr}
\toprule
Model & Valid F1 \\
\midrule 
\textit{Wizard of Wikipedia} & \\ 
Previous Utterance Only & 24.6  \\ 
        + dialog Context & 26.4 \\ 
        + Turn Embedding & 27.4 \\ 
\midrule 
\textit{Engaging ImageChat} & \\ 
Previous Utterance Only & 13.3  \\ 
        + dialog Context & 14.5 \\ 
        + Turn Embedding + Personality & 15.1 \\ 
\bottomrule
\end{tabular}
\caption{\textbf{Important Features for KNN Search} using KIF. Salient conversation features improve performance on both datasets.}
\label{tab:encoding_features_quality}
\end{table}

\begin{table}
\centering
\begin{tabular}[t]{lr}
\toprule
Model & Valid F1 \\
\midrule
KIF-Augmented Transformer & 27.4 \\ 
\midrule 
\multicolumn{2}{c}{One KIF Module fetches multiple times} \\ 
2 Fetches & 26.9 \\ 
3 Fetches & 26.0 \\ 
\midrule 
\multicolumn{2}{c}{Multiple KIF Modules fetch once each} \\
2 Fetches & 26.5\\ 
3 Fetches & 25.9\\ 
\bottomrule
\end{tabular}
\caption{\textbf{Multi-hop with KIF} to retrieve information with multiple fetch steps.}
\label{tab:multihop}
\end{table}

\paragraph{Using dialog Features for KNN Performance.}

The quality of the KNN search is critical  to the performance of KIF modules. As the external knowledge is kept fixed, KIF must be able to align the dialog context with the knowledge to identify relevant pieces of information. In Table~\ref{tab:encoding_features_quality}, we show that matching on more features can improve  the  quality  of the retrieved information.  Using only the encoding of the immediate previous utterance can improve results on Wizard of Wikipedia by 7 F1 points, but this is further improved by also leveraging the encoding of context (+1.8 F1) and using the dialog turn number (+1 F1).  These features are available in the datasets, and we leverage them to improve the relatedness of retrieved knowledge. 

\paragraph{Multi-Hop Retrieval with KIF.} Work in memory networks \cite{weston2014memory, sukhbaatar2015end} employed multi-hop mechanisms. Such capacity could be useful when multiple sources are necessary or information is incrementally fetched. To emulate multi-hop memory mechanisms, we use KIF to retrieve relevant information for $N = 2$ or $N = 3$ fixed hops. As the number of hops is fixed, the multi-hop operation remains differentiable. We do not allow the model to retrieve the same information in a second hop.

We experimented in two settings. First, the same KIF module is used multiple times to fetch different information, and then all of the fetched knowledge is concatenated. Results are shown in Table~\ref{tab:multihop} (top). Second, we examine spreading the fetches into \textit{different} KIF modules at various encoder depths. This could be interpreted as the model learning to access more information each layer. As the model progresses deeper, more abstract and high level representations are built, which could allow different knowledge to be retrieved. Results are shown in Table~\ref{tab:multihop} (bottom). 

In both multi-hop settings, no improvement in performance on the Wizard of Wikipedia dataset is observed. We hypothesize this can be partially attributed to the construction of the dataset --- as humans explicitly based their written dialog utterance on one knowledge sentence. Further, it is possible that concatenation brings together too much information for the model to incorporate, and thus adding additional fetches makes the retrieval more noisy.

\paragraph{Effect of Gating.}

We analyze the effect of the gating mechanism by evaluating the capability of the gate to identify and focus on salient information. 
On Wizard of Wikipedia, we concatenate a third source of information: dialog turns from a completely different corpus called PersonaChat \citep{zhang2018personalizing}. This dataset looks quite different --- short utterances without factual knowledge ---  and should be easy for the model to identify as distinct from Wizard of Wikipedia.  As shown in Figure~\ref{fig:ablation_1}(b), if KIF on PersonaChat is included without gating, it has a harmful effect as the model includes irrelevant information. When equipped with gating, the model learns to use the gate to ignore some inputs, and can recover almost the full performance of a model without this irrelevant information source.

\paragraph{Size of K in KNN.}

Figure~\ref{fig:ablation_1}(c) shows the performance on Wizard of Wikipedia when varying the amount of knowledge. Being able to access multiple relevant pieces of information is helpful, but too much information can be harmful. This is likely because the weighted sum becomes  blurry if too many sentences are incorporated.

\section{Conclusion}

We present a KNN-based Information Fetching module that learns to identify relevant information from external knowledge sources by learning a mapping-based read operation. KIF modules benefit from the scalability and efficiency of K Nearest Neighbors search, enabling computation with large external memories. We show in the context of two dialog datasets that relevant knowledge can be identified and incorporated to create more engaging, high quality dialog.

\section*{Acknowledgements}

We thank the reviewers and action editor for their comments and insightful discussion. We thank Emily Dinan and Kurt Shuster for providing assistance to reproduce their original works. 

\bibliography{tacl2018}
\bibliographystyle{acl_natbib}

\end{document}

%% file: math_commands.tex
\usepackage{amsmath,amsfonts,bm}

\def\eqref#1{equation~\ref{#1}}

\def\1{\bm{1}}

\DeclareMathAlphabet{\mathsfit}{\encodingdefault}{\sfdefault}{m}{sl}
\SetMathAlphabet{\mathsfit}{bold}{\encodingdefault}{\sfdefault}{bx}{n}

\newcommand{\sigmoid}{\sigma}

